\documentclass{llncs} %
\pdfoutput=1
\usepackage{hyperref}
\usepackage{url}
\usepackage{graphicx}
\usepackage{tikz}

\usepackage{amsmath}

\title{Linear State-Space Model with\\ Time-Varying Dynamics}

\author{
Jaakko Luttinen \and Tapani Raiko \and Alexander Ilin
}

\institute{Aalto University, Finland}

\newcommand{\mx}[1]{\mathbf{#1}}
\newcommand{\mxg}[1]{\boldsymbol{#1}}
\newcommand{\T}{^{\operatorname{T}}}

\newcommand{\inv}{^{-1}}

\DeclareMathOperator{\diag}{diag}

\newcommand{\obsm}{\mathcal{O}_{m:}}

\begin{document}

\maketitle

\begin{abstract}
  This paper introduces a linear state-space model with time-varying dynamics.
  The time dependency is obtained by forming the state dynamics matrix as a
  time-varying linear combination of a set of matrices.  The time dependency of
  the weights in the linear combination is modelled by another linear Gaussian
  dynamical model allowing the model to learn how the dynamics of the process
  changes.  Previous approaches have used switching models which have a small
  set of possible state dynamics matrices and the model selects one of
  those matrices at each time, thus jumping between them.  Our model forms the
  dynamics as a linear combination and the changes can be smooth and more
  continuous.  The model is motivated by physical processes which are described
  by linear partial differential equations whose parameters vary in time.  An
  example of such a process could be a temperature field whose evolution is
  driven by a varying wind direction.  The posterior inference is performed
  using variational Bayesian approximation.  The experiments on stochastic
  advection-diffusion processes and real-world weather processes show that the
  model with time-varying dynamics can outperform previously introduced
  approaches.
\end{abstract}

\section{Introduction}

Linear state-space models (LSSM) are widely used in time-series analysis and
modelling of dynamical systems \cite{Bar-Shalom:2001,Shumway:2000}.  They assume
that the observations are generated linearly from hidden states with a linear
dynamical model that does not change with time. The assumptions of linearity and
constant dynamics make the model easy to analyze and efficient to learn.

Most real-world processes cannot be accurately described by linear Gaussian
models, which motivates more complex nonlinear state-space models (see, e.g.,
\cite{Ghahramani:1999,valpola2002}).  However, in many cases processes behave
approximately linearly in a local regime.  For instance, an industrial process
may have a set of regimes with very distinct but linear dynamics.  Such
processes can be modelled by switching linear state-space models
\cite{Ghahramani:1998:sssm,Pavlovic:2000} in which the transition between a set
of linear dynamical models is described with hidden Markov models. Thus, these
models have a small number of states defining their dynamics.

Instead of having a small number of possible states of the process dynamics,
some processes may have linear dynamics that change continuously in time.  For
instance, physical processes may be characterized by linear stochastic partial
differential equations but the parameters of the equations may vary in time.
Simple climate models may use the advection-diffusion equation in which the
diffusion and the velocity field parameters define how the modelled quantity
mixes and moves in space and time.  In a realistic scenario, these parameters
are time-dependent because, for instance, the wind modelled by the velocity
field changes with time.

This paper presents a Bayesian linear state-space model with time-varying
dynamics.  The dynamics at each time is formed as a linear combination of a set
of state dynamics matrices, and the weights of the linear combination follow a
linear Gaussian dynamical model.  The main difference to switching LSSMs is that
instead of having a small number of dynamical regimes, the proposed model allows
for an infinite number of them with a smooth transition between them.  Thus, the
model can adapt to small changes in the system.  This work is an extension of an
abstract \cite{raiko2010drifting} which presented the basic idea without the
Bayesian treatment.  The model bears some similarity to relational feature
learning in modelling sequential data \cite{michalski2014}.

Posterior inference for the model is performed using variational Bayesian (VB)
approximation because the exact Bayesian inference is intractable
\cite{Beal:2003}.  In order for the VB learning algorithm to converge fast, the
method uses a similar parameter expansion that was introduced in
\cite{Luttinen:2013}.  This parameter expansion is based on finding the optimal
rotation in the latent subspace and it may improve the speed of the convergence
by several orders of magnitude.

The experimental section shows that the proposed LSSM with time-varying dynamics
is able to learn the varying dynamics of complex physical processes.  The model
predicts the processes better than the classical LSSM and the LSSM with switching
dynamics.  It finds latent processes that describe the changes in the dynamics
and is thus able to learn the dynamics at each time point accurately.  These
experimental results are promising and suggest that the time-varying dynamics
may be a useful tool for statistical modelling of complex dynamical and physical
processes.

\section{Model}
\label{sec:model}

Linear state-space models assume that a sequence of $M$-dimensional observations
$(\mx{y}_1, \ldots, \mx{y}_N)$ is generated from latent $D$-dimensional states
$(\mx{x}_1, \ldots, \mx{x}_N)$ following a first-order Gaussian Markov process:
\begin{align}
  \mathbf{y}_n &= \mathbf{C} \mathbf{x}_n + \mathrm{noise},
  \\
  \mathbf{x}_n &= \mathbf{W} \mathbf{x}_{n-1} + \mathrm{noise},
  \label{eq:static_x}
\end{align}
where noise is Gaussian, $\mx{W}$ is the $D\times D$ state dynamics matrix and
$\mx{C}$ is the $M \times D$ loading matrix.  Usually, the latent space
dimensionality $D$ is assumed to be much smaller than the observation space
dimensionality $M$ in order to model the dependencies of high-dimensional
observations efficiently.  Because the state dynamics matrix is constant, the
model can perform badly if the dynamics of the modelled process changes in time.

In order to model changing dynamics, the constant dynamics in
\eqref{eq:static_x} can be replaced with a state dynamics matrix $\mx{W}_n$
which is time-dependent.  Thus, \eqref{eq:static_x} is replaced with
\begin{align}
  \mathbf{x}_n &= \mathbf{W}_n \mathbf{x}_{n-1} + \mathrm{noise}.
\end{align}
However, modelling the unknown time dependency of $\mathbf{W}_n$ is a
challenging task because for each $\mx{W}_n$ there is only one transition
$\mx{x}_{n-1} \rightarrow \mx{x}_n$ which gives information about each
$\mathbf{W}_n$.%

Previous work modelled the time-dependency using switching state dynamics
\cite{Pavlovic:2000}.  It means having a small set of matrices $\mathbf{B}_1,
\ldots, \mathbf{B}_K$ and using one of them at each time step:
\begin{align}
  \mathbf{W}_n &= \mathbf{B}_{z_n}, \label{eq:switching_W}
\end{align}
where $z_n\in \{1,\ldots,K\}$ is a time-dependent index.  The indices $z_n$ then
follow a first-order Markov chain with an unknown state-transition matrix.  The
model can be motivated by dynamical processes which have a few states with
different dynamics and the process jumps between these states.

This paper presents an approach for continuously changing time-dependent
dynamics.  The state dynamics matrix is constructed as a linear combination of
$K$ matrices:
\begin{align}
  \mathbf{W}_n &= \sum^K_{k=1} s_{kn} \mathbf{B}_k . \label{eq:varying_W}
\end{align}
The mixing weight vector $\mx{s}_n=\begin{bmatrix} s_{1n} & \ldots &
  s_{Kn} \end{bmatrix}\T$ varies in time and follows a first-order Gaussian
Markov process:
\begin{align}
  \mathbf{s}_n &= \mathbf{A} \mathbf{s}_{n-1} + \mathrm{noise},
\end{align}
where $\mathbf{A}$ is the $K\times{}K$ state dynamics matrix of this latent
mixing-weight process.  The model with switching dynamics in
\eqref{eq:switching_W} can be interpreted as a special case of
\eqref{eq:varying_W} by restricting the weight vector $\mx{s}_n$ to be a binary
vector with only one non-zero element.  However, in the switching model,
$\mx{s}_n$ would follow a first-order Markov chain, which is different from the
first-order Gaussian Markov process used in the proposed model.  Compared to
models with switching dynamics, the model with time-varying dynamics allows the
state dynamics matrix to change continuously and smoothly.

The model is motivated by physical processes which roughly follow stochastic
partial differential equations but the parameters of the equations change in
time.  For instance, a temperature field may be modelled with a stochastic
advection-diffusion equation but the direction of the wind may change in time,
thus changing the velocity field parameter of the equation.

\subsection{Prior Probability Distributions}

We give the proposed model a Bayesian formulation by setting prior probability
distributions for the variables.  It roughly follows the linear state-space
model formulation in \cite{Beal:2003,Luttinen:2013} and the principal component
analysis formulation in \cite{Bishop:1999}.  The likelihood function is
\begin{align}
  p(\mathbf{Y}|\mathbf{C},\mathbf{X},\tau) &= \prod^N_{n=1}
  \mathcal{N}(\mathbf{y}_n | \mathbf{C}\mathbf{x}_n,
  \diag(\boldsymbol\tau)\inv),
\end{align}
where $\mathcal{N}(y|m,v)$ is the Gaussian probability density function of $y$
with mean $m$ and covariance $v$, and $\diag(\boldsymbol\tau)$ is a diagonal
matrix with elements $\tau_1, \ldots, \tau_M$ on the diagonal.  For simplicity,
we used isotropic noise ($\tau_m=\tau$) in our experiments.

The loading matrix $\mx{C}$ has the following prior, which is also known as an
automatic relevance determination (ARD) prior \cite{Bishop:1999}:
\begin{align}
  p(\mx{C}|\mxg{\gamma}) &= \prod^D_{d=1} \mathcal{N} ( \mx{c}_d | \mx{0},
  \diag(\mxg{\gamma})\inv ),
  &
  p(\mxg{\gamma}) &= \prod^D_{d=1} \mathcal{G}( \gamma_d | a_\gamma, b_\gamma),
\end{align}
where $\mx{c}_d$ is the $d$-th row vector of $\mx{C}$, the vector $\mxg{\gamma}
= \begin{bmatrix} \gamma_1 & \ldots & \gamma_D \end{bmatrix}\T$ contains the ARD
parameters, and $\mathcal{G}( \gamma | a, b)$ is the gamma probability density
function of $\gamma$ with shape $a$ and rate $b$.

The latent states $\mx{X}=\begin{bmatrix} \mx{x}_0 & \ldots &
  \mx{x}_N \end{bmatrix}$ follow a first-order Gaussian Markov process, which
can be written as
\begin{align}
  p(\mathbf{X} | \mathbf{W}_n) &= \mathcal{N}(\mx{x}_0|\mxg{\mu}_0^{(x)},
  \mx{\Lambda}_0\inv) \prod^N_{n=1} \mathcal{N}(\mx{x}_n | \mx{W}_n
  \mx{x}_{n-1}, \mx{I}),
\end{align}
where $\mxg{\mu}_0^{(x)}$ and $\mx{\Lambda}_0$ are the mean and precision of the
auxiliary initial state $\mx{x}_0$.  The process noise covariance matrix can be
an identity matrix without loss of generality because any rotation can be
compensated in $\mx{x}_n$ and $\mx{W}_n$.  The initial state $\mx{x}_0$ can be
given a broad prior by setting, for instance, $\mxg{\mu}_0^{(x)}=\mx{0}$ and
$\mx{\Lambda}_0=10^{-6}\cdot\mx{I}$.

The state dynamics matrices $\mx{W}_n$ are a linear combination of matrices
$\mx{B}_k$ which have the following ARD prior:
\begin{align}
  p(\mathbf{B}_k|\boldsymbol{\beta}_k) &= \prod^D_{c=1} \prod^D_{d=1}
  \mathcal{N}(b_{kcd} | \mx{0}, \beta_{kd}\inv),
  &
  \!\!\!\! p(\beta_{dk}) &= \mathcal{G} (\beta_{kd}| a_\beta, b_\beta),
  &
  \!\!\!\! k &= 1, \ldots, K,
\end{align}
where $b_{kcd}=[\mx{B}_k]_{cd}$ is the element on the $c$-th row and $d$-th
column of $\mx{B}_k$.  The ARD parameter $\beta_{kd}$ helps in pruning out
irrelevant components in each matrix.

In order to keep the formulas less cluttered, we use the following notation:
$\mx{B}$ is a $K\times{}D\times{}D$ tensor.  When using subscripts, the first
index corresponds to the index of the state dynamics matrix, the second index to
the rows of the matrices and the third index to the columns of the matrices.  A
colon is used to denote that all elements along that axis are taken.  Thus, for
instance, $\mx{B}_{k::}$ is $\mx{B}_k$ and $\mx{B}_{:d:}$ is a $K\times{}D$
matrix obtained by stacking the $d$-th row vectors of $\mx{B}_k$ for each $k$.

The mixing weights $\mx{S}=\begin{bmatrix} \mx{s}_0 & \ldots &
  \mx{s}_N \end{bmatrix}$ have first-order Gaussian Markov process prior
\begin{align}
  p(\mathbf{S}|\mx{A}) &= \mathcal{N}(\mx{s}_0 | \mxg{\mu}^{(s)}_0,
  \mx{V}_0\inv) \prod^N_{n=1} \mathcal{N} (\mx{s}_n | \mx{A}\mx{s}_{n-1}, \mx{I}),
\end{align}
where, similarly to the prior of $\mx{X}$, the parameters $\mxg{\mu}^{(s)}_0$
and $\mx{V}_0$ are the mean and precision of the auxiliary initial state
$\mx{s}_0$, and the noise covariance can be an identity matrix without loss of
generality.  The initial state $\mx{s}_0$ can be given a broad prior by setting,
for instance, $\mxg{\mu}^{(s)}_0=\mx{0}$ and $\mx{V}_0=10^{-6}\cdot\mx{I}$.

The state dynamics matrix $\mx{A}$ of the latent mixing weights $\mx{s}_n$ is
given an ARD prior
\begin{align}
  p(\mx{A}|\mxg{\alpha}) &= \prod^K_{k=1} \mathcal{N} \left( \mx{a}_k | \mx{0},
    \diag(\mxg{\alpha})\inv \right),
  &
  p(\mxg{\alpha}) &= \prod^K_{k=1} \mathcal{G}( \alpha_k | a_\alpha, b_\alpha)
\end{align}
where $\mx{a}_k$ is the $k$-th row of $\mx{A}$, and $\mxg{\alpha}=\begin{bmatrix}
  \alpha_1 & \ldots & \alpha_K \end{bmatrix}\T$ contains the ARD parameters.

Finally, the noise parameter is given a gamma prior
\begin{align}
  p(\boldsymbol{\tau}) &= \prod^M_{m=1} \mathcal{G}(\tau_m | a_\tau, b_\tau).
\end{align}
The hyperparameters of the model can be set, for instance, as $a_\alpha =
b_\alpha = a_\beta = b_\beta = a_\gamma = b_\gamma = a_\tau = b_\tau = 10^{-6}$
to obtain broad priors for the variables.  Small values result in approximately
non-informative priors which are usually a good choice in a wide range of
problems.

\begin{figure}[tb]
\usetikzlibrary{shapes}
\usetikzlibrary{fit}
\usetikzlibrary{chains}
\usetikzlibrary{arrows}
\usetikzlibrary{bayesnet}
\centering
  \begin{tikzpicture}

       \tikzstyle{latent} += [minimum size=20pt];
       
       \node[latent] (x0) {$\mathbf{x}_0$};
       \node[latent, below right=of x0] (x1) {$\mathbf{x}_1$};
       \node[right=of x1] (dots) {$\cdots$};
       \node[latent, right=of dots] (xn) {$\mathbf{x}_N$};
       \edge {x0}{x1};
       \edge {x1}{dots};
       \edge {dots}{xn};

       \node[obs, below=1 of x1] (y1) {$\mathbf{y}_1$};
       \node[right=of y1] (ydots) {$\cdots$};
       \node[obs, right=of ydots] (yn) {$\mathbf{y}_N$};
       \edge {x1} {y1} ;
       \edge {dots} {ydots} ;
       \edge {xn} {yn} ;

       \node[latent, above=1 of x1] (s1) {$\mathbf{s}_1$};
       \node[latent, above left=of s1] (s0) {$\mathbf{s}_0$};
       \node[right=of s1] (sdots) {$\cdots$};
       \node[latent, right=of sdots] (sn) {$\mathbf{s}_N$};

       \node[latent, above right=of xn] (B) {$\mathbf{B}$};
       \node[latent, right=1 of B] (beta) {$\boldsymbol{\beta}$};
       \edge {beta} {B} ;
       
       \edge {B} {x1, dots, xn};
       \edge {s0} {s1} ;
       \edge {s1} {x1};
       \edge {s1} {sdots};
       \edge {sdots} {dots};
       \edge {sdots} {sn} ;
       \edge {sn} {xn};

       \node[latent, above right=of sn] (A) {$\mathbf{A}$} ;
       \edge {A} {s1,sdots,sn};
       \node[latent, right=1 of A] (alpha) {$\boldsymbol{\alpha}$};
       \edge {alpha} {A} ;

       \node[latent, above right=1 of yn] (C) {$\mathbf{C}$} ;
       \node[latent, right=1 of C] (gamma) {$\boldsymbol{\gamma}$};
       \edge {C} {y1,ydots,yn};
       \edge {gamma} {C} ;

       \node[latent, above left=1 of y1] (tau) {$\boldsymbol{\tau}$};
       \edge {tau} {y1,ydots,yn} ;

  \end{tikzpicture} 
  \caption{The graphical model of the linear state-space model with time-varying
    dynamics}
\end{figure}
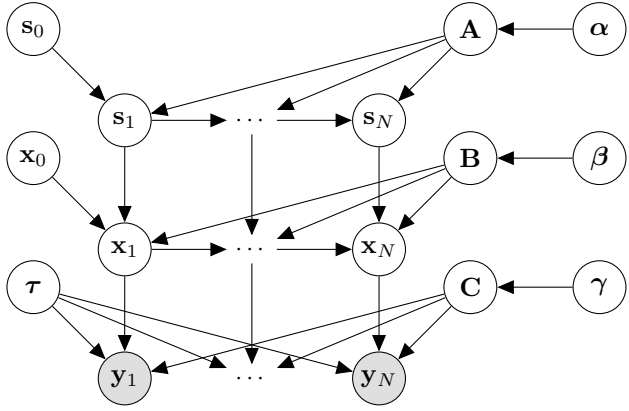

For the experimental section, we constructed the LSSM with switching dynamics by
using a hidden Markov model (HMM) for the state dynamics matrix $\mx{W}_n$.  The
HMM had an unknown initial state and a state transition matrix with broad
conjugate priors.  We used similar prior probability distributions in the
classical LSSM with constant dynamics, the proposed LSSM with time-varying
dynamics, and the LSSM with switching dynamics for the similar parts of the
models.

\section{Variational Bayesian Inference}
\label{sec:inference}

As the posterior distribution is analytically intractable, it is approximated
using variational Bayesian (VB) framework, which scales well to large
applications compared to Markov chain Monte Carlo (MCMC) methods
\cite{Bishop:2006}.  The posterior approximation is assumed to factorize with
respect to the variables:
\begin{align}
  p(\mx{X},\!\mx{C},\!\mxg{\gamma},\mx{B},\mxg{\beta},\mx{S},\mx{A},\mxg{\alpha},\mxg{\tau}|\mx{Y})
  \!\approx\!
  q(\mx{X})q(\mx{C})q(\mxg{\gamma})q(\mx{B})q(\mxg{\beta})q(\mx{S})
  q(\mx{A})q(\mxg{\alpha})q(\mxg{\tau}).
\end{align}
This approximation is optimized by minimizing the Kullback-Leibler divergence
from the true posterior
by using the variational Bayesian expectation-maximization (VB-EM) algorithm
\cite{Beal:2003b}.  In VB-EM, the posterior approximation is updated for the
variables one at a time and iterated until convergence.

\subsection{Update Equations}

The approximate posterior distributions have the following forms:
\begin{align}
  q(\mx{X}) &= \mathcal{N}([\mx{X}]_:|\mxg{\mu}_{x}, \mx{\Sigma}_x),
  &
  q(\mxg{\tau}) &= \prod^M_{m=1} \mathcal{G} (\tau_m | \bar{a}_\tau^{(m)},
  \bar{b}_\tau^{(m)}),
  \\
  q(\mx{C}) &= \prod^M_{m=1} \mathcal{N}( \mx{c}_{m} | \mxg{\mu}_c^{(m)}, \mx{\Sigma}_c^{(m)}),
  &
  q(\mxg{\gamma}) &= \prod^D_{d=1} \mathcal{G} (\gamma_d| \bar{a}_\gamma^{(d)},
  \bar{b}_\gamma^{(d)}),
  \\
  q(\mx{B}) &= \prod^D_{d=1} \mathcal{N} ( [\mx{B}_{:d:}]_: | \mxg{\mu}_b^{(d)},
  \mx{\Sigma}_b^{(d)}),
  &
  q(\mxg{\beta}) &= \prod^K_{k=1} \prod^D_{d=1} \mathcal{G} ( \beta_{kd} |
  \bar{a}_\beta^{(kd)}, \bar{b}_\beta^{(kd)}),
  \\
  q(\mx{S}) &= \mathcal{N}( [\mx{S}]_: | \mxg{\mu}_s, \mx{\Sigma}_s ),
  \\
  q(\mx{A}) &= \prod^K_{k=1} \mathcal{N}(\mx{a}_{k} | \mxg{\mu}_a^{(k)}, \mx{\Sigma}_a^{(k)}),
  &
  q(\mxg{\alpha}) &= \prod^K_{k=1} \mathcal{G} (\alpha_k| \bar{a}_\alpha^{(k)},
  \bar{b}_\alpha^{(k)}),
\end{align}
where $[\mx{X}]_:$ is a vector obtained by stacking the column vectors
$\mx{x}_n$.  It is straightforward to derive the following update equations of
the variational parameters:
\begin{align}
  \bar{a}_\tau^{(m)} &= a_\tau + \frac{N_m}{2},
  &
  \bar{b}_\tau^{(m)} &= b_\tau + \frac{1}{2} \sum_{n\in\obsm} \xi_{mn},
  \\
  \mx{\Sigma}_c^{(m)} &= \left( \! \langle\diag(\mxg{\gamma})\rangle +
    \!\!\! \sum_{n\in\obsm} \!\!\langle\tau_m\rangle \langle\mx{x}_n\mx{x}_n\T\rangle
  \!\right)\inv  \!\!\!\!\!\!\!,
  &
  \mxg{\mu}_c^{(m)} &= \mx{\Sigma}_c^{(m)} \!\!\!\sum_{n\in\obsm} \!\!\! y_{mn}
  \langle{\tau_m}\rangle \langle\mx{x}_n\rangle,
  \\
  \bar{a}_\gamma^{(d)} &= a_\gamma + \frac{M}{2},
  &
  \bar{b}_\gamma^{(d)} &= b_\gamma + \frac{1}{2} \sum^M_{m=1} \langle
  c_{md}^2\rangle,
\end{align}
\begin{align}
  \mx{\Sigma}_b^{(d)} &= \left( \langle\diag(\mxg{\beta})\rangle +
    \sum^{N}_{n=1} \mx{\Omega}_n \right)\inv \!\!\!\!\!\!\!,
  &
  \mxg{\mu}_b^{(d)} &= \mx{\Sigma}_b^{(d)} \!\sum^N_{n=1} \!\left[ \langle
  \mx{s}_{n}\rangle \langle x_{dn}\mx{x}_{n-1}\T \rangle \right]_: ,
  \\
  \mx{\Sigma}_a^{(k)} &= \left( \! \langle\diag(\mxg{\alpha})\rangle +
    \!\sum^{N}_{n=1} \langle\mx{s}_{n-1}\mx{s}_{n-1}\T\rangle \!\right)\inv
  \!\!\!\!\!\!\!,
  &
  \mxg{\mu}_a^{(k)} &= \mx{\Sigma}_a^{(k)} \sum^N_{n=1} \langle
  s_{kn}\mx{s}_{n-1}\rangle,
  \\
  \bar{a}_\alpha^{(k)} &= a_\alpha + \frac{K}{2},
  &
  \bar{b}_\alpha^{(k)} &= b_\alpha + \frac{1}{2} \sum^K_{i=1} \langle a_{ik}^2
  \rangle,
\end{align}
where $\obsm$ is the set of time instances $n$ for which the observation
$y_{mn}$ is not missing, $N_m$ is the size of the set $\obsm$, $\xi_{mn} =
\left\langle (y_{mn} - \mx{c}_m\T\mx{x}_n)^2 \right\rangle$, $\mx{\Omega}_n =
\langle\mx{x}_{n-1}\mx{x}_{n-1}\T\rangle \otimes
\langle\mx{s}_n\mx{s}_n\T\rangle$, and $\otimes$ denotes the Kronecker product.
The computation of the posterior distribution of $\mx{X}$ and $\mx{S}$ is more
complicated and will be discussed next.

The time-series variables $\mx{X}$ and $\mx{S}$ can be updated using algorithms
similar to the Kalman filter and the Rauch-Tung-Striebel smoother.  The
classical formulations of those algorithms do not work for VB learning because
of the uncertainty in the dynamics matrix \cite{Beal:2003,Barber:2007}.  Thus,
we used a modified version of these algorithms as presented for the classical
LSSM in \cite{Luttinen:2013}.  The algorithm performs a forward and a backward
pass in order to find the required posterior expectations.

The explicit update equations for $q(\mx{X})$ can be written as:
\begin{align}
  \mx{\Sigma}_x\inv &=
  \arraycolsep=6.0pt
  \def\arraystretch{1.5}
  \begin{bmatrix}
    \mx{\Lambda}_0 + \langle \mx{W}_1\T\mx{W}_1 \rangle &
    -\langle \mx{W}_1 \rangle\T &
    &
    \\
    -\langle \mx{W}_1 \rangle &
    \mx{I} +  \langle \mx{W}_2\T\mx{W}_2 \rangle + \mx{\Psi}_1 &
    \ddots
    &
    \\
    &
    \ddots &
    \ddots &
    -\langle \mx{W}_N \rangle\T
    \\
    &
    &
    -\langle \mx{W}_N \rangle &
    \mx{I} + \mx{\Psi}_N
  \end{bmatrix},
\end{align}
\begin{align}
  \mxg{\mu}_x &= \mx{\Sigma}_x
  \def\arraystretch{1.5}
  \begin{bmatrix}
    \mx{\Lambda}_0 \mxg{\mu}_0^{(x)}
    \\
    \sum_{m \in \mathcal{O}_{:1}} y_{m1} \langle\tau_m\rangle
    \langle\mx{c}_m\rangle
    \\
    \vdots
    \\
    \sum_{m \in \mathcal{O}_{:N}} y_{mN} \langle\tau_m\rangle
    \langle\mx{c}_m\rangle
  \end{bmatrix},
\end{align}
where $\mathcal{O}_{:n}$ is the set of indices $m$ for which the observation
$y_{mn}$ is not missing, $\mx{\Psi}_n = \sum_{m \in \mathcal{O}_{:n}}
\langle\tau_m\rangle \langle\mx{c}_m\mx{c}_m\T\rangle$,
$\langle\mx{W}_n\rangle=\sum_{k=1}^K\langle s_{kn} \rangle
\langle\mx{B}_k\rangle$, and $\langle\mx{W}_n\T\mx{W}_n\rangle =
\sum_{k=1}^K\sum_{l=1}^K [\langle \mx{s}_{n}\mx{s}_{n}\T \rangle]_{kl}
\langle\mx{B}_k\T\mx{B}_l\rangle$.  Instead of using standard matrix inversion,
one can utilize the block-banded structure of $\mx{\Sigma}_x\inv$ to compute the
required expectations $\langle\mx{x}_n\rangle$,
$\langle\mx{x}_n\mx{x}_n\T\rangle$ and $\langle\mx{x}_n\mx{x}_{n-1}\T\rangle$
efficiently.  The algorithm for the computations is presented in
\cite{Luttinen:2013}.

Similarly for $\mx{S}$, the explicit update equations are
\begin{align}
  \mx{\Sigma}_s\inv &=
  \arraycolsep=6.0pt
  \def\arraystretch{1.5}
  \begin{bmatrix}
    \mx{V}_0 + \langle \mx{A}\T\mx{A} \rangle &
    -\langle \mx{A} \rangle\T &
    &
    \\
    -\langle \mx{A} \rangle &
    \mx{I} + \langle \mx{A}\T\mx{A} \rangle + \mx{\Theta}_1 &
    \ddots
    &
    \\
    &
    \ddots &
    \ddots &
    -\langle \mx{A} \rangle\T
    \\
    &
    &
    -\langle \mx{A} \rangle &
    \mx{I} + \mx{\Theta}_N
  \end{bmatrix},
\end{align}
\begin{align}
  \mxg{\mu}_s &= \mx{\Sigma}_s
  \def\arraystretch{1.5}
  \begin{bmatrix}
    \mx{V}_0 \mxg{\mu}_0^{(s)}
    \\
    \sum_{d=1}^D \langle\mx{B}_{:d:}\rangle \langle x_{d1} \mx{x}_0\T \rangle
    \\
    \vdots
    \\
    \sum_{d=1}^D \langle\mx{B}_{:d:}\rangle \langle x_{dN} \mx{x}_{N-1}\T \rangle
  \end{bmatrix},
\end{align}
where $\mx{\Theta}_n = \sum_{i=1}^D\sum_{j=1}^D
[\langle\mx{x}_{n-1}\mx{x}_{n-1}\T\rangle]_{ij} \cdot
\langle\mx{B}_{::i}\mx{B}_{::j}\T\rangle$.  The required expectations
$\langle\mx{s}_n\rangle$, $\langle\mx{s}_n\mx{s}_n\rangle$ and
$\langle\mx{s}_n\mx{s}_{n-1}\rangle$ can be computed efficiently by using the
same algorithm as for $\mx{X}$ \cite{Luttinen:2013}.

The VB learning of the LSSM with switching dynamics is quite similar to the
equations presented above.  The main difference is that the posterior
distribution of the discrete state variable $z_n$ is computed by using
alpha-beta recursion \cite{Bishop:2006}.  The update equations for the state
transition probability matrix and the initial state probabilities are
straightforward because of the conjugacy.  The expectations
$\langle\mx{W}_n\rangle$ and $\langle\mx{W}_n\T\mx{W}_n\rangle$ are computed by
averaging $\langle\mx{B}_k\rangle$ and $\langle\mx{B}_k\T\mx{B}_k\rangle$ over
the state probabilities $\mathrm{E}[z_n=k]$.

\subsection{Practical Issues}

The main practical issue with the proposed model is that the VB learning
algorithm may converge to bad local minima.  As a solution, we found two ways of
improving the robustness of the method.  The first improvement is related to the
updating of the posterior approximation and the second improvement is related to
the initialization of the approximate posterior distributions.

The first practical tip is that one may want to run the VB updates for the lower
layers of the model hierarchy first for a few times before starting to update
the upper layers.  Otherwise, the hyperparameters may learn very bad values
because the child variables have not yet been well estimated.  Thus, we updated
$\mx{X}$, $\mx{C}$, $\mx{B}$ and $\mxg{\tau}$ 5--10 times before updating the
hyperparameters and the upper layers.  However, this procedure requires a
reasonable initialization.

We initialized $\mx{X}$ and $\mx{C}$ randomly but for $\mx{S}$ and $\mx{B}$ we
used a bit more complicated approach.  One goal of the initialization was that
the model would be close to a model with constant dynamics.  Thus, the first
component in $\mx{S}$ was set to a constant value and the corresponding matrix
$\mx{B}_k$ was initialized as an identity matrix.  The other components in
$\mx{S}$ and $\mx{B}$ were random but their scale was a bit smaller so that the
time variation in the resulting state dynamics matrix $\mx{W}_n$ was small
initially.  Obviously, this initialization leads to a bias towards a constant
component in $\mx{S}$ but this is often realistic as the system probably has
some average dynamics and deviations from it.

\subsection{Rotations for Faster Convergence}

One issue with the VB-EM algorithm for state-space models is that the algorithm
may converge extremely slowly.  This happens if the variables are strongly
correlated because they are updated only one at a time causing zigzagging and
small updates.  This effect can be reduced by the parameter expansion approach,
which finds a suitable auxiliary parameter connecting several variables and then
optimizes this auxiliary parameter \cite{Liu:1998,Qi:2007}.  This corresponds to
a parameterized joint optimization of several variables.

A suitable parameter expansion for state-space models is related to the rotation
of the latent sub-space \cite{Luttinen:2010,Luttinen:2013}.  It can be motivated
by noting that the latent variable $\mx{X}$ can be rotated arbitrarily by
compensating it in $\mx{C}$:
\begin{align}
  \mx{y}_n &= \mx{C}\mx{x}_n = \mx{C}\mx{R}\inv\mx{R}\mx{x}_n =
  \big(\mx{C}\mx{R}\inv\big) \big(\mx{R}\mx{x}_n\big) \quad \text{ for all
    non-singular } \mx{R} \,.
\end{align}
The rotation of $\mx{X}$ must also be compensated in the dynamics $\mx{W}_n$ as
\begin{align}
  \mx{R}\mx{x}_n &= \mx{R}\mx{W}_n\mx{R}\inv \mx{R}\mx{x}_{n-1} =
  \big(\mx{R}\mx{W}_n\mx{R}\inv\big) \big(\mx{R}\mx{x}_{n-1}\big) \,.
\end{align}
The rotation $\mx{R}$ can be used to parameterize the posterior distributions
$q(\mx{X})$, $q(\mx{C})$ and $q(\mx{B})$.  Optionally, the distributions of the
hyperparameters $q(\mxg{\gamma})$ and $q(\mxg{\beta})$ can also be
parameterized.  Optimizing the posterior approximation with respect to $\mx{R}$
is efficient and leads to significant improvement in the speed of the VB
learning.  Details for the procedure in the context of the classical LSSM can be
found in \cite{Luttinen:2013}.

Similarly to $\mx{X}$, the latent mixing weights $\mx{S}$ can also be rotated as
\begin{align}
  [\mx{W}_n]_{d:} &= \mx{B}_{:d:}\T \mx{s}_n = \mx{B}_{:d:}\T\mx{R}\inv\mx{R}\mx{s}_n
  = \Big(\mx{B}_{:d:}\T\mx{R}\inv\Big) \Big(\mx{R}\mx{s}_n\Big) \,,
\end{align}
where $[\mx{W}_n]_{d:}$ is the $d$-th row vector of $\mx{W}_n$.  The rotation
must also be compensated in the dynamics of $\mx{S}$ as
\begin{align}
  \mx{R}\mx{s}_n &= \mx{R}\mx{A}\mx{R}\inv \mx{R}\mx{s}_{n-1} =
  \big(\mx{R}\mx{A}\mx{R}\inv\big) \big(\mx{R}\mx{s}_{n-1}\big) \,.
\end{align}
Thus, the rotation corresponds to a parameterized joint optimization of
$q(\mx{S})$, $q(\mx{A})$, $q(\mx{B})$, and optionally also $q(\mxg{\alpha})$ and
$q(\mxg{\beta})$.  Note that the optimal rotation of $\mx{S}$ can be computed
separately from the optimal rotation for $\mx{X}$.

The extra computational cost by the rotation speed up is small compared to the
computational cost of one VB update of all variables.  Thus, the rotation can be
computed at each iteration after the variables have been updated.  If for some
reason the computation of the optimal rotation is slow, one can use the
rotations less frequently, for instance, after every ten updates of all
variables, and still gain similar performance improvements.  However, as was
shown in \cite{Luttinen:2013}, the rotation transformation is essential even for
the classical LSSM, thus ignoring it may lead to extremely slow convergence and
poor results.  Thus, we used the rotation transformation for all methods in the
next section.

\section{Experiments}
\label{sec:experiments}

We compare the proposed linear state-space model with time-varying dynamics
(LSSM-TVD) to the classical linear-state space model (LSSM) and the linear
state-space model with switching dynamics (LSSM-SD) using three datasets: a
one-dimensional signal with changing frequency, a simulated physical process
with time-varying parameters, and real-world daily temperature measurements in
Europe.  The methods are evaluated by their ability to predict missing values
and gaps in the observed processes.

\subsection{Signal with Changing Frequency}

\begin{figure}[tb]
  \small
  \begin{center}
    \begin{tabular}{c}
      \includegraphics[width=\linewidth]{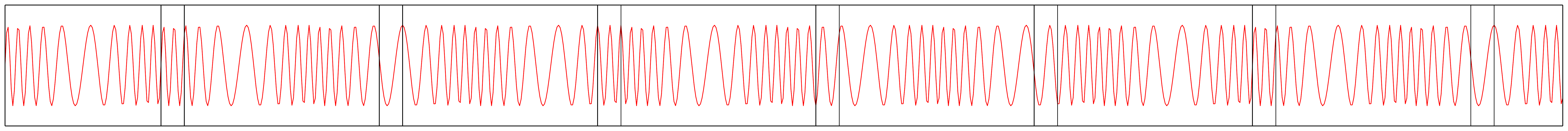}
      \\
      (a) True signal
      \\
      \includegraphics[width=\linewidth]{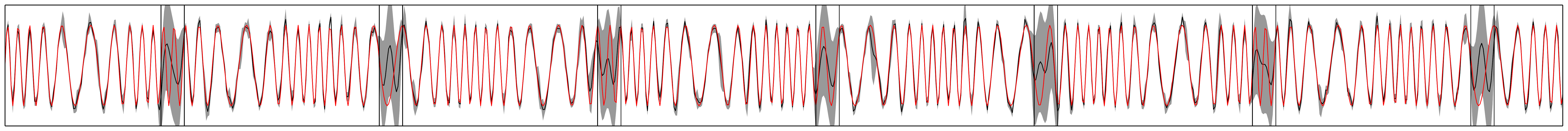}
      \\
      (b) LSSM
      \\
      \includegraphics[width=\linewidth]{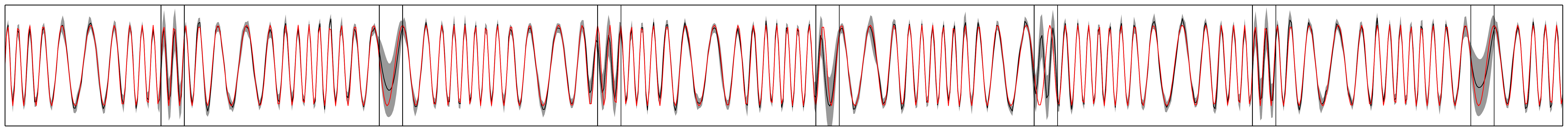}
      \\
      (c) LSSM-SD
      \\
      \includegraphics[width=\linewidth]{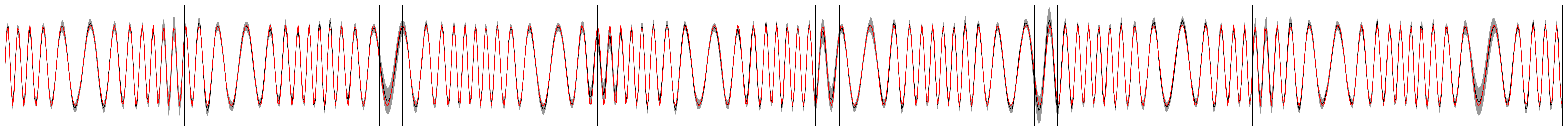}
      \\
      (d) LSSM-TVD
    \end{tabular}
  \end{center}
  \caption{Results for the signal with changing frequency: (a) the true signal,
    (b) the classical LSSM, (c) the LSSM with switching dynamics, (d) the LSSM
    with time-varying dynamics.  In (b)-(d), the posterior mean is shown as
    solid black line, two standard deviations are shown as a gray area, and the
    true signal is shown in red for comparison.  Vertical lines mark the seven
    gaps that contain no observations.}
  \label{fig:artificial}
\end{figure}

We demonstrate the LSSM with time-varying dynamics using an artificial signal
with changing frequency.  The signal is defined as
\begin{align}
  f(n) = \sin ( a \cdot (n + c \sin(b \cdot 2\pi) ) \cdot 2\pi ), \quad
  n=0,\ldots,999
\end{align}
where $a=0.1$, $b=0.01$ and $c=8$.  The resulting signal is shown in
Fig.~\ref{fig:artificial}(a).  The signal was corrupted with Gaussian noise
having zero mean and standard deviation $0.1$ to simulate noisy observations.
In order to see how well the different methods can learn the dynamics, we
created seven gaps in the signal by removing 15 consecutive observations to
produce each gap.  In addition, 20\% of the remaining observations were randomly
removed.  Each method (LSSM, LSSM-SD and LSSM-TVD) used $D=5$ dimensions for the
latent states $\mx{x}_n$.  The LSSM-SD and LSSM-TVD used $K=4$ state dynamics
matrices $\mx{B}_k$.

Figures~\ref{fig:artificial}(b)-(d) show the posterior distribution of the
latent noiseless function $f$ for each method. The classical LSSM is not able to
capture the dynamics and the reconstructions over the gaps are bad and have high
variance. The LSSM-SD learns two different states for the dynamics corresponding
to a lower and a higher frequency.  The reconstructions over the gaps are better
than with the LSSM, but it still has quite a large variance and the fifth gap is
reconstructed using a wrong frequency.  The gap reconstructions have large
variance because the two state dynamics matrices learned by the model do not fit
the process very well so the model assumes a larger innovation noise in the
latent process $\mx{X}$. In contrast to that, the LSSM-TVD learns the dynamics
practically perfectly and even learns the dynamics of the process which changes
the frequency.  Thus, the LSSM-TVD is able to make nearly perfect predictions
over the gaps and the variance is small.  It also prunes out one state dynamics
matrix, thus using effectively only three dimensions for the latent
mixing-weight process.

\subsection{Stochastic Advection-Diffusion Process}

We simulated a physical process with time-dependent parameters in order to
compare the considered approaches.  The physical process is a stochastic
advection-diffusion process, which is defined by the following partial
differential equation:
\begin{align}
  \frac{ \partial f}{ \partial t } &= \delta \nabla^2 f - \mathbf{v} \cdot
  \nabla f + R ,
  \label{eq:spde}
\end{align}
where $f$ is the variable of interest, $\delta$ is the diffusivity, $\mathbf{v}$
is the velocity field and $R$ is a stochastic source.  We have assumed that the
diffusivity is a constant and the velocity field describes an incompressible
flow.  The velocity field $\mathbf{v}$ changes in time.  This equation could
describe, for instance, air temperature and the velocity field corresponds to
winds with changing directions.  The spatial domain was a torus, a
two-dimensional manifold with periodic boundary conditions.

The partial differential equation \eqref{eq:spde} is discretized using the finite
difference method.  This is used to generate a discretized realization of the
stochastic process by iterating over the time domain.  The stochastic source $R$
is a realization from a spatial Gaussian process at each time step.  The two
velocity field components are modelled as follows:
\begin{align}
  \mx{v}(t+1) &= \sqrt{\rho}\cdot \mx{v}(t) + \sqrt{1-\rho} \cdot
  \boldsymbol{\xi}(t+1),
\end{align}
where $\rho\in (0,1)$ controls how fast the velocity changes and
$\boldsymbol{\xi}(t+1)$ is Gaussian noise with zero mean and variance which was
chosen appropriately.  Thus, there are actually two sources of randomness in the
stochastic process: the random source $R$ and the randomly changing velocity
field $\mathbf{v}$.

The data were generated from the simulated process as follows: Every 20-th
sample was kept in the time domain, which resulted in $N=2000$ time instances.
From the spatial discretization grid, $M=100$ locations were selected randomly
as the measurement locations (corresponding to weather stations).  The simulated
values were corrupted with Gaussian noise to obtain noisy observations.

We used four methods in this comparison: LSSM, LSSM-SD and LSSM-TVD with $D=30$
dimensions for the latent states $\mx{x}_n$, and LSSM with $D=60$ to see if
adding more dimensions improves the performance of the classical LSSM.  Both the
LSSM-SD and LSSM-TVD used $K=5$ state dynamics matrices $\mx{B}_k$.

\begin{figure}[tb]
  \centering
  \includegraphics[width=0.35\linewidth]{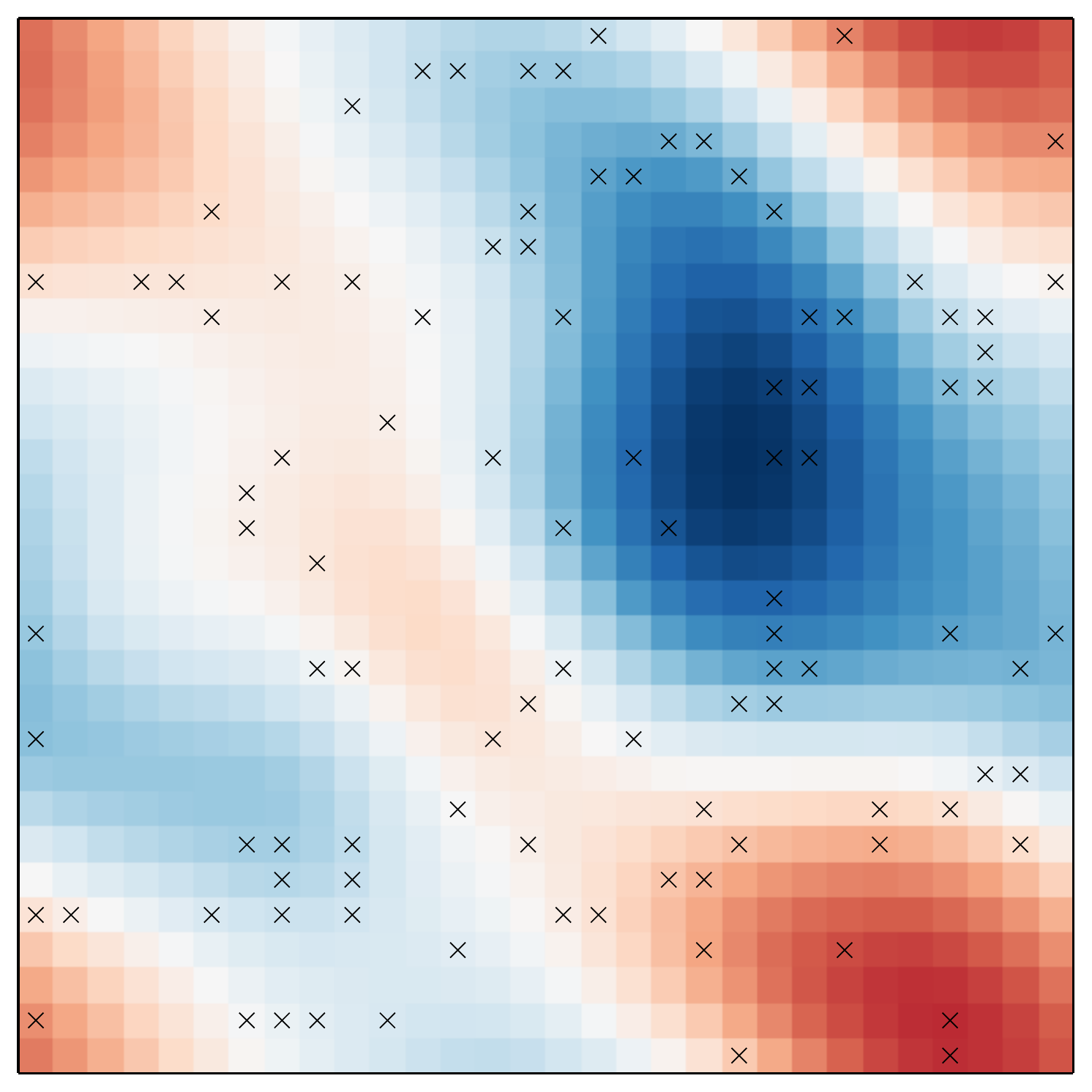}
  \caption{One of the simulated processes at one time instance.  Crosses denote the
    locations that were used to collect the observations.  Note that the domain
    is a torus, that is, a 2-dimensional manifold with periodic boundaries.}
  \label{fig:spde}
\end{figure}

For measuring the performance of the methods, we generated two test sets.
First, we created 18 gaps of 15 consecutive time points, that is, the
observations from all the spatial locations were removed over the gaps and the
corresponding values of the noiseless process $f$ formed the first test set.
Second, we randomly removed 20\% of the remaining observations and used the
corresponding values of the process $f$ as the second test set.  The tests were
performed for five simulated processes.  Figure~\ref{fig:spde} shows one process
at one time instance as an
example.\footnote{\url{http://users.ics.aalto.fi/jluttine/ecml2014/} contains a
  video visualization of each of the simulated processes.}

Table~\ref{tab:spde} shows the root-mean-square errors (RMSE) of the mean
reconstructions for both the generated gaps and the randomly removed values.
The results for each of the five process realizations are shown separately.  It
appears that using $D=60$ components does not significantly change the
performance of the LSSM compared to using $D=30$ components.  Also, the LSSM-SD
performs practically identically to the LSSM. The LSSM-SD used effectively two
or three state dynamics matrices.  However, this does not seem to help in
modelling the variations in the dynamics and the learned model performs
similarly to the LSSM. In contrast to that, the proposed LSSM-TVD has the best
performance in each experiment.  For the test set of random values, the
difference is not large because the reconstruction is mainly based on the
correlations between the locations rather than the dynamics.  However, in order
to accurately reconstruct the gaps, the model needs to learn the changes in the
dynamics. The LSSM-TVD reconstructs the gaps more accurately than the other
methods, because it finds latent mixing weights $\mx{s}_n$ which model the
changes in the dynamics.

\begin{table}[b]
  \centering
  \caption{Results for five stochastic advection-diffusion experiments.  The
    root-mean-square errors (RMSE) have been multiplied by a factor of 1000 for
    clarity.}
  
  \small

  \setlength{\tabcolsep}{6.5pt}
  \begin{tabular}{c|ccccc|ccccc}
    &
    \multicolumn{5}{c|}{RMSE for gaps} &
    \multicolumn{5}{c}{RMSE for random}
    \\
    Method &   1 & 2 & 3 & 4 & 5   &    1 & 2 & 3 & 4 & 5
    \\
    \hline
    LSSM $D=30$  &
    104 & 107 & 102 & 94 & 104 &
    34 & 38 & 39 & 34 & 34
    \\
    LSSM $D=60$  &
    105 & 107 & 110 & 98 & 108 &
    35 & 39 & 40 & 35 & 35
    \\
    LSSM-SD $D=30$ &
    106 & 117 & 113 & 94 & 102 &
    35 & 37 & 39 & 34 & 34 
    \\
    LSSM-TVD $D=30$ &
    \bf 73 & \bf 81 & \bf 75 & \bf 67 & \bf 82 &
    \bf 30 & \bf 34 & \bf 35 & \bf 31 & \bf 30
  \end{tabular}
  \label{tab:spde}
\end{table}

\begin{figure}[tb]
  \centering
  \begin{tabular}{cc}
    \includegraphics[width=0.49\linewidth]{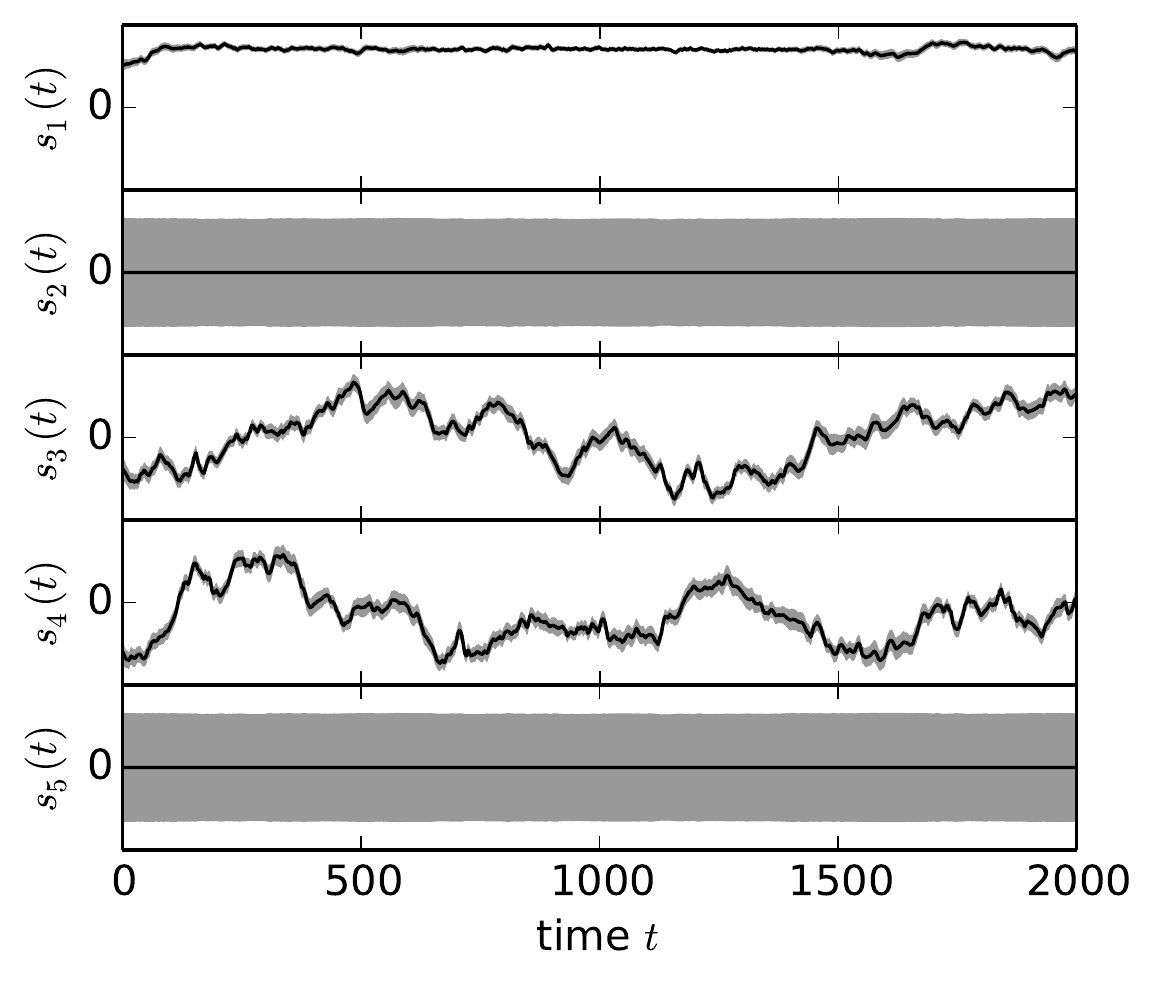}
    &
    \includegraphics[width=0.49\linewidth]{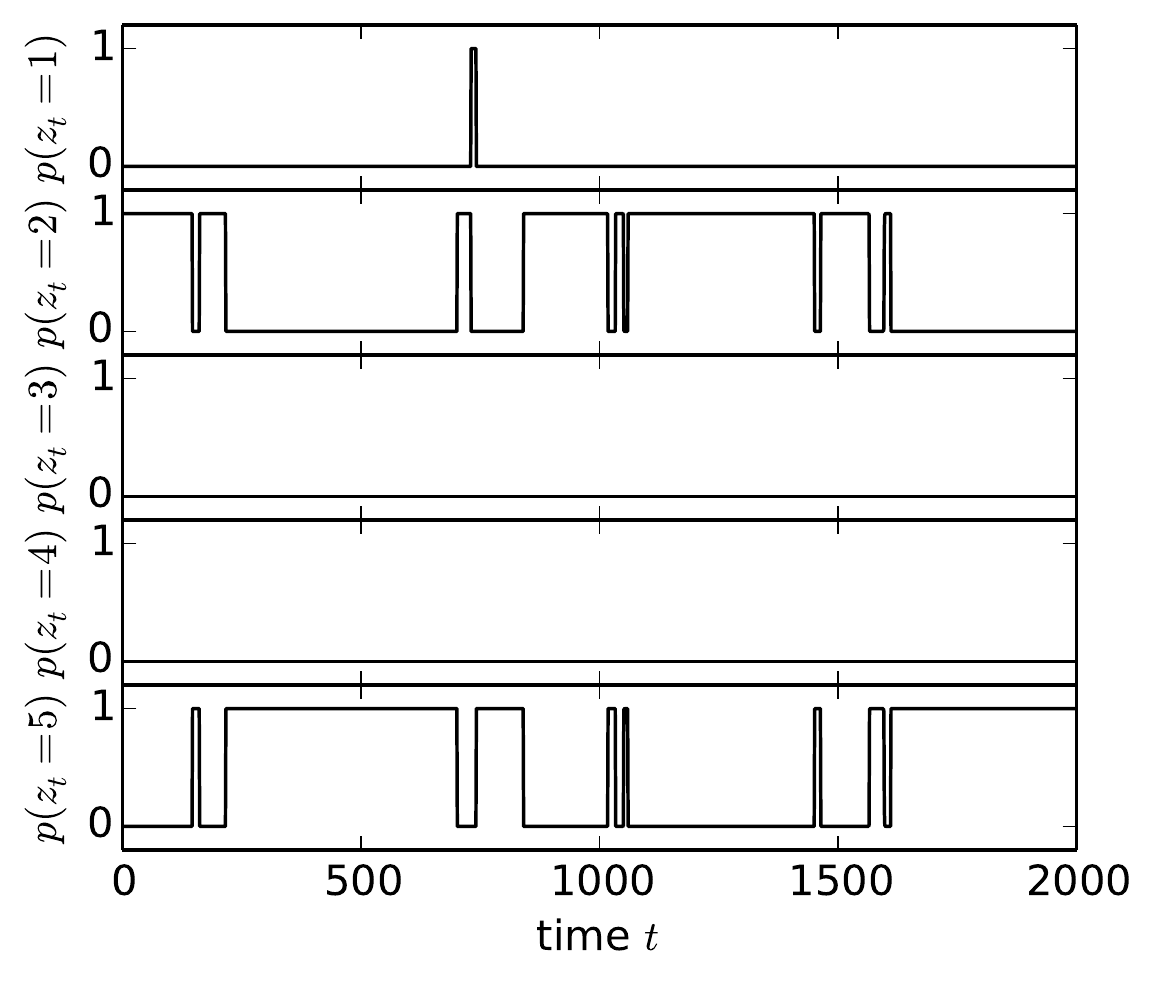}
    \\
    (a) $q(\mathbf{S})$ in LSSM-TVD
    &
    (b) $q(\mathbf{Z})$ in LSSM-SD
  \end{tabular}
  \caption{Results for the advection-diffusion experiments.  (a) The posterior
    mean and two standard deviations of the latent mixing weights by the
    LSSM-TVD.  (b) The posterior probability of each state transition matrix as
    a function of time in the LSSM-SD.}
  \label{fig:spde_S}
\end{figure}

Figure~\ref{fig:spde_S}(a) shows the the posterior distribution of the $K=5$
mixing-weight signals $\mx{S}$ in one experiment: the first signal is
practically constant corresponding to the average dynamics, the third and the
fourth signals correspond to the changes in the two-dimensional velocity field,
and the second and the fifth signals have been pruned out as they are not
needed.  Thus, the method was able to learn the effective dimensionality of the
latent mixing-weight process, which suggests that the method is not very
sensitive to the choice of $K$ as long as it is large enough.  The results look
similar in all the experiments with the LSSM-TVD and in every experiment the
LSSM-TVD found one constant and two varying components.  Thus, the posterior
distribution of $\mx{S}$ might give insight on some latent processes that affect
the dynamics of the observed process.

This experiment showed that the LSSM-SD is not good at modelling linear
combinations of the state dynamics matrices.  Interestingly, although the VB
update formulas average the state dynamics matrices by their probabilities
resulting in a convex combination,
this mixing is not very prominent in the approximate posterior distribution as
seen in Fig.~\ref{fig:spde_S}(b).
Most of the time, only one state dynamics matrix is active with probability one.
This happens because the prior penalizes switching between the matrices and one
of the matrices is usually much better than the others on average over several
time steps.

\subsection{Daily Mean Temperature}

The third experiment used real-world temperature measurements in Europe.  The
data were taken from the global surface summary of day product produced by the
National Climatic Data Center (NCDC) \cite{NCDC}.  We studied daily mean
temperature measurements roughly from the European area\footnote{The longitude
  of the studied region was in range $(-13, 33)$ and the latitude in range $(35,
  72)$.} in 2000--2009.  Stations that had more than 20\% of the measurements
missing were discarded.  This resulted in $N=3653$ time instances and $M=1669$
stations for the analysis.

The three models were learned from the data.  They used $D=80$ dimensions for
the latent states.  The LSSM-SD and the LSSM-TVD used $K=6$ state dynamics
matrices.  We formed two test sets similarly to the previous experiment.  First,
we generated randomly 300 2-day gaps in the data, which means that measurements
from all the stations were removed during those periods of time.  Second, 20\%
of the remaining data was used randomly to form another test set.

Table~\ref{tab:gsod} shows the results for five experiments using different test
sets.  The methods reconstructed the randomly formed test sets equally well
suggesting that learning more complex dynamics did not help and the learned
correlations between the stations was sufficient.  However, the reconstruction
of gaps is more interesting because it measures how well the method learns the
dynamical structure.  This reconstruction shows consistent performance
differences between the methods: The LSSM-SD is slightly worse than the LSSM,
and the LSSM-TVD outperforms the other two.  Because climate is a chaotic
process, the modelling is extremely challenging and predictions tend to be far
from perfect.  However, these results suggest that the time-varying dynamics
might offer a promising improvement to the classical LSSM in statistical
modelling of physical processes.

\begin{table}[tb]
  \centering
  \caption{GSOD reconstruction errors of the test sets in degrees Celsius for five
    runs }
  \label{tab:gsod}
  \small

  \setlength{\tabcolsep}{3.1pt}
  \begin{tabular}{c|ccccc|ccccc}
    &
    \multicolumn{5}{c|}{RMSE for gaps} &
    \multicolumn{5}{c}{RMSE for randomly missing}
    \\
    Method &   1 & 2 & 3 & 4 & 5   &    1 & 2 & 3 & 4 & 5
    \\
    \hline
    LSSM   &
    1.748 & 1.753 & 1.758 & 1.744 & 1.751 &   
    0.935 & 0.937 & 0.935 & 0.933 & 0.934
    \\
    LSSM-SD  &
    1.800 & 1.801 & 1.796 & 1.777 & 1.788 & 
    0.936 & 0.938 & 0.936 & 0.934 & 0.935
    \\
    LSSM-TVD  &
    \bf 1.661 & \bf 1.650 & \bf 1.659 & \bf 1.653 & \bf 1.660 &   
    0.935 & 0.937 & 0.935 & 0.932 & 0.934
  \end{tabular}
\end{table}

\section{Conclusions}
\label{sec:conclusions}

This paper introduced a linear state-space model with time-varying dynamics.  It
forms the state dynamics matrix as a time-varying linear combination of a set of
matrices.  It uses another linear state-space model for the mixing weights in
the linear combination.  This is different from previous time-dependent LSSMs
which use switching models to jump between a small set of states defining the
model dynamics.

Both the LSSM with switching dynamics and the proposed LSSM are useful but they
are suitable for slightly different problems.  The switching dynamics is
realistic for processes which have a few possible states that can be quite
different from each other but each of them has approximately linear dynamics.
The proposed model, on the other hand, is realistic when the dynamics vary more
freely and continuously.  It was largely motivated by physical processes based
on stochastic partial differential equations with time-varying parameters.

The experiments showed that the proposed LSSM with time-varying dynamics can
capture changes in the underlying dynamics of complex processes and
significantly improve over the classical LSSM.  If these changes are continuous
rather than discrete jumps between a few states, it may achieve better modelling
performance than the LSSM with switching dynamics.  The experiment on a
stochastic advection-diffusion process showed how the proposed model adapts to
the current dynamics at each time and finds the current velocity field which
defines the dynamics.

The proposed model could be further improved for challenging real-world
spatio-temporal modelling problems.  First, the spatial structure could be taken
into account in the prior of the loading matrix using, for instance, Gaussian
processes \cite{Luttinen:2009:nips}.  Second, outliers and badly corrupted
measurements could be modelled by replacing the Gaussian observation noise
distribution with a more heavy-tailed distribution, such as the Student-$t$
distribution \cite{Luttinen:2012:npl}.

The method was implemented in Python as a module for an open-source variational
Bayesian package called BayesPy \cite{BayesPy}.  It is distributed under an open
license, thus making it easy for others to apply the method.  In addition, the
scripts for reproducing all the experimental results are also
available.\footnote{Experiment scripts available at
  \url{http://users.ics.aalto.fi/jluttine/ecml2014/}}

\subsubsection*{Acknowledgments.}

We would like to thank Harri Valpola, Natalia Korsakova, and Erkki Oja for
useful discussions. This work has been supported by the Academy of Finland
(project number 134935).

\bibliographystyle{splncs}
\bibliography{bibliography}

\end{document}